\documentclass[letterpaper, 10 pt, conference]{ieeeconf} 

\IEEEoverridecommandlockouts                              

\usepackage{amsmath,amssymb}
\usepackage{times}      
\usepackage{mathptmx}   

\usepackage{graphicx}   
\usepackage{booktabs}
\usepackage{array}
\usepackage{tabularx}
\usepackage{makecell}
\usepackage{ragged2e}
\usepackage{threeparttable}
\usepackage{afterpage}

\usepackage[ruled,vlined,linesnumbered,noend]{algorithm2e}
\DontPrintSemicolon
\SetKwInput{KwInput}{Inputs}
\SetKwInput{KwOutput}{Outputs}
\SetKwInput{KwData}{State}

\let\labelindent\relax
\usepackage[inline]{enumitem} 

\usepackage[hidelinks]{hyperref}
\usepackage{xcolor}

\newcolumntype{Y}{>{\RaggedRight\arraybackslash}X}
\newcolumntype{P}[1]{>{\RaggedRight\arraybackslash}p{#1}}
\renewcommand{\arraystretch}{1.1}

\title{\LARGE \bf
AIRHILT: A Human-in-the-Loop Testbed for Multimodal Conflict Detection in Aviation
}

\author{Omar Garib$^{1}$, Jayaprakash D. Kambhampaty$^{1}$, Olivia J. Pinon Fischer$^{1}$, and Dimitri N. Mavris$^{1}$
\thanks{$^{1}$Daniel Guggenheim School of Aerospace Engineering, Georgia Institute of Technology}%
}

\begin{document}

\maketitle
\thispagestyle{empty}
\pagestyle{empty}

\begin{abstract} We introduce AIRHILT (Aviation Integrated Reasoning, Human-in-the-Loop Testbed), a modular and lightweight simulation environment designed to evaluate multimodal pilot and ATC assistance systems for aviation conflict detection. Built on the Godot engine~\cite{godot_engine_godot_nodate}, AIRHILT synchronizes pilot and air traffic controller (ATC) communications, visual scene understanding from camera streams, and ADS-B surveillance data within a unified, scalable platform. The environment supports pilot- and controller-in-the-loop interactions, providing a comprehensive scenario suite covering both terminal area and en route operational conflicts, including communication errors and procedural mistakes. AIRHILT offers standardized JSON-based interfaces enabling researchers to easily integrate, swap, and evaluate various automatic speech recognition (ASR), visual detection, decision-making, and text-to-speech (TTS) models. We demonstrate AIRHILT through a reference pipeline incorporating fine-tuned Whisper ASR, YOLO-based visual detection, ADS-B-based conflict logic, and GPT-OSS-20B structured reasoning, presenting preliminary results from representative runway-overlap scenarios where the assistant achieves an average time-to-first-warning of ${\sim}7.7$\,s with average ASR and vision latencies of ${\sim}5.9$\,s and ${\sim}0.4$\,s, respectively. The AIRHILT environment and scenario suite are openly available, supporting reproducible research on multimodal situational awareness and conflict detection in aviation. The complete repository is available at \href{https://github.com/ogarib3/airhilt}{\texttt{github.com/ogarib3/airhilt}}.
\end{abstract}

\section{INTRODUCTION}
Aircraft situational awareness and conflict detection currently rely on accurate multi-aircraft surveillance and radio communications, with command and control of airspace operations managed centrally by human air traffic controllers (ATCs). However, human ATCs and pilots are vulnerable to overwork, fatigue, and loss of attention, increasing the risk of operational errors and conflict events \cite{della_rocco_role_1999, rosekind_assessing_2024}. 

To alleviate workload pressures and reduce operational errors, there is a growing need for assistive aviation systems that incorporate recent advancements in \emph{automatic speech recognition} (ASR) and \emph{vision-based detection}, alongside existing aircraft and radar surveillance data (e.g., ADS-B, radar). Such systems could proactively identify hazards such as traffic conflicts and runway incursions, while enhancing situational awareness and minimizing additional workload for pilots and controllers.

\begin{figure}[t!]
  \centering
  \setlength{\fboxsep}{0pt}%
  \setlength{\fboxrule}{1.2pt}%

  \fbox{\includegraphics[width=\linewidth]{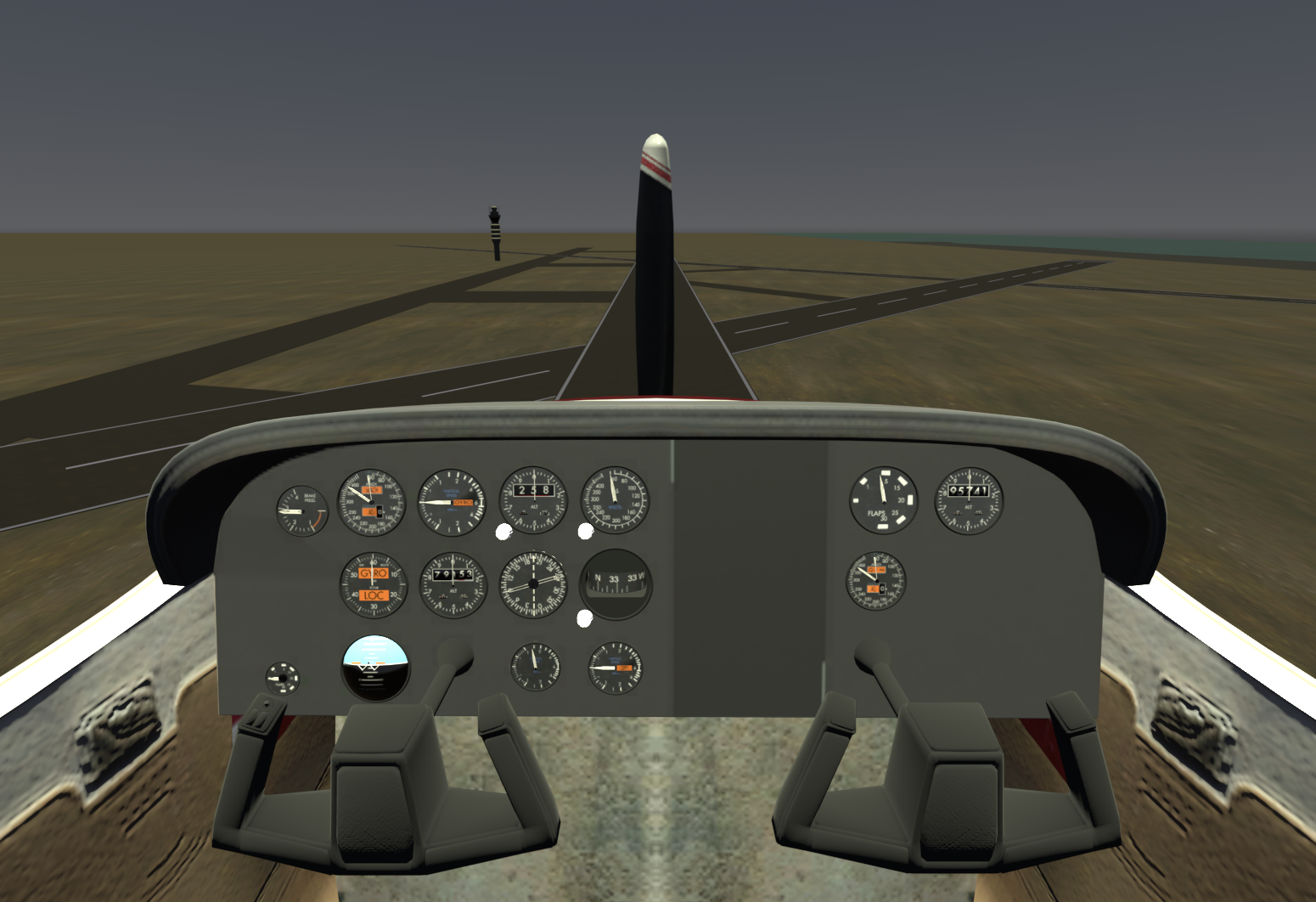}}

  \caption{\textbf{AIRHILT at a glance.} Cockpit view from the simulation showing the setting used for pilot‑in‑the‑loop evaluations. AIRHILT synchronizes radio traffic, vision feeds, and ADS‑B to test end‑to‑end assistive warning pipelines.}
  \label{fig:teaser}
\end{figure}

However, the current testing landscape for such aviation assistive systems presents notable challenges. Physically collocated test environments are costly, time-consuming, and require careful scheduling of limited ATC and pilot availability. These sophisticated facilities typically include several pilot and controller workstations along with integrated displays and are commonly utilized for operational scenario studies~\cite{schier_designing_2013, manske_visual_2015}. Yet, their availability is increasingly constrained by growing global aviation traffic and rising research demands associated with emerging aviation concepts such as unmanned aerial systems (UAS) and advanced air mobility (AAM)~\cite{chaisit_enhancing_2024, kaliardos_identifying_2024}. Although rigorous testing at physical facilities remains essential for advanced development and certification phases, it is impractical for early-stage concept evaluations. Additionally, the rapidly expanding design space, driven by new machine learning models and varied computational, sensing, and communication architectures, further underscores the necessity of flexible, efficient, simulation-based environments suitable for rapid, systematic evaluations.

Such a simulation-based environment would significantly broaden access to aviation situational awareness research, allowing researchers worldwide to efficiently explore and identify promising candidate systems while conserving limited pilot and ATC resources. To address these needs, we introduce \emph{AIRHILT}, a simulation environment explicitly designed to facilitate research into multimodal AI assistance systems through pilot and controller-in-the-loop experimentation.

\textbf{Contributions.} Our contributions in this effort are as follows:
\begin{enumerate}
\item \textbf{An open simulation environment} that synchronizes pilot-to-ATC communications, ATC control tower camera views, aircraft-mounted camera streams, and ADS-B/radar data, enabling systematic evaluation of multimodal assistive systems with pilot- and controller-in-the-loop interactions.

\item \textbf{A scalable scenario suite} consisting of six conflict scenario families (three terminal and three en route) that model communication, procedural, and visually driven hazards, with parameterized variations in noise, visibility, geometry, and traffic configurations.

\item \textbf{A reference multimodal pipeline} that demonstrates environment capabilities through interchangeable components such as Whisper-based ASR, YOLO-based visual detection, ADS-B-based conflict logic, and a structured large language model (LLM) decision layer, with preliminary latency and time-to-first-warning metrics reported.

\item \textbf{Reproducible artifacts} including the simulation environment, scenario definitions, evaluation scripts, and documentation to support community use and extension.
\end{enumerate}

The remainder of this paper is structured as follows: Section~\ref{sec:background} provides background on air traffic management operations and the key components relevant to aviation situational awareness. Section~\ref{sec:challenges} outlines related challenges to building such systems. Section~\ref{sec:env} details the environment and interfaces. Section~\ref{sec:scenarios} presents the designed scenarios. Section~\ref{sec:reference_pipeline} describes the reference pipeline and presents preliminary results from representative runway-overlap scenarios. Section~\ref{sec:limitations} provides concluding remarks, discusses current limitations, and introduces avenues for future work.

\section{Background} \label{sec:background}

Aviation operations encompass both air traffic management and conflict detection, including the determination, sequencing, and issuance of clearances from departure through en route and approach phases \cite{international_civil_aviation_organization_procedures_2016, federal_aviation_administration_air_2024}, as well as the identification and mitigation of hazards such as wildlife encounters, mechanical issues, and other unexpected conflicts \cite{federal_aviation_administration_hazardous_nodate}. Effective traffic management maintains prescribed separation between aircraft while preserving operational efficiency under current operational conditions. Controllers integrate surveillance data (e.g., ADS-B and radar) provided via tower infrastructure, direct visual observations from the control tower, and standardized voice communications with pilots to formulate and issue clearances and instructions~\cite{federal_aviation_administration_acas_nodate}. In parallel, pilots routinely manage additional hazards, including wildlife (e.g., birds) activity, uncooperative or untracked intruder aircraft, and onboard mechanical anomalies, by synthesizing external visual cues with onboard sensor data and established procedures.

Many unsafe conditions arise from (i) \emph{traffic-management breakdowns} (e.g., violations of separation minima or runway occupancy conflicts) and (ii) \emph{aircraft or airspace hazards} (e.g., bird strikes, uncooperative intruder aircraft, or mechanical anomalies). It is important to note that existing radar-based surveillance systems such as Automatic Dependent Surveillance-Broadcast (ADS-B), and collision-avoidance systems such as the Traffic Collision Avoidance System (TCAS/ACAS), already provide critical support. However, incidents have occurred and continue to occur even with these systems in place. For example, the Überlingen mid-air collision highlighted critical vulnerabilities when ATC instructions conflict with TCAS resolution advisories (RAs), reinforcing that TCAS RAs must always take precedence over controller clearances \cite{international_air_transport_association_acas_nodate}. Moreover, it is important to note that TCAS RAs are intentionally inhibited at low altitudes \cite{federal_aviation_administration_acas_nodate}, impacting its operation to resolve situations such as runway incursions. These realities motivate the development of an assistive layer that fuses visual streams, voice communications, and surveillance data to identify and flag potential conflicts \emph{before} situations escalate to require intervention from safety systems such as TCAS.

\subsection{Operational Communication Loop} We focus on the standard pilot–controller communication loop used across terminal and en route operations. At a high level: \begin{enumerate}
\item Pilot \emph{establishes contact with the appropriate ATC facility} (e.g., tower or approach).

\item Controller \emph{issues an instruction or clearance}.

\item Pilot \emph{reads back} the instruction or clearance.

\item Controller \emph{monitors} the readback, correcting if necessary, after which the pilot executes. \end{enumerate} 

Failures can occur at multiple points (e.g., mishearing, incorrect readback, or delayed compliance), motivating assistive monitoring across modalities.

\subsection{Multimodal Components and Perception Tasks}
Recent progress in multimodal AI provides essential building blocks for assistance: (i) \textbf{ASR} for transcribing and parsing ATC/pilot communications; (ii) \textbf{vision detection models} operating on tower or onboard cameras for surface/aircraft/vehicle detection and tracking; and (iii) \textbf{surveillance and trajectory analytics} from ADS-B/radar data. However, performance thresholds and latency budgets required for operational usefulness remain under-specified, and real systems must handle missing or degraded modalities.

\begin{figure*}[!t]
  \centering
  \includegraphics[width=\textwidth,keepaspectratio]{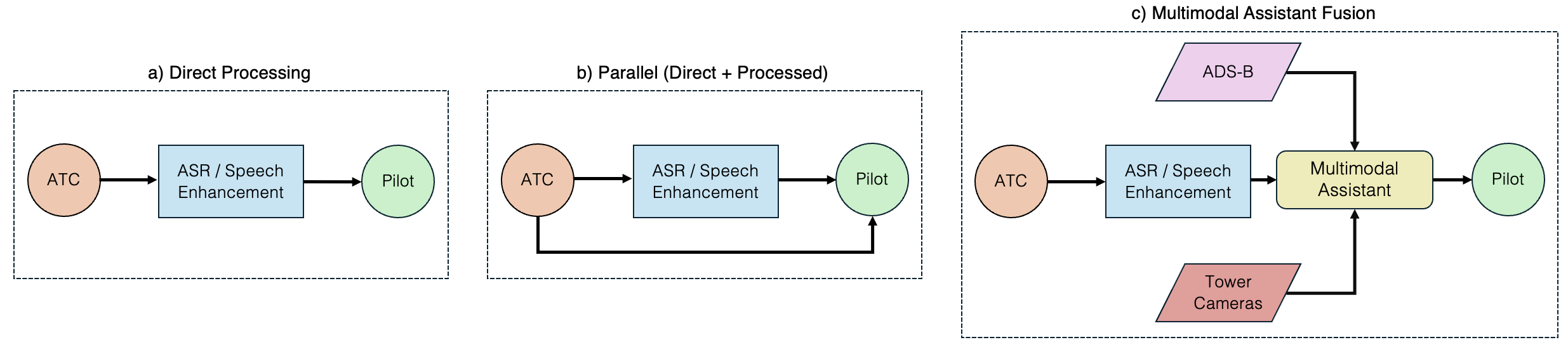}

  \caption{Canonical topology alternatives for assistive processing in the ATC--pilot loop.
(a) \textit{ASR/SE first}: audio is enhanced (SE) and/or transcribed (ASR) first; the resulting output directly informs the advisory presented to the pilot;
(b) \textit{Parallel paths}: raw audio reaches the pilot while a copy is processed by ASR/SE in parallel;
(c) \textit{Assistant-gated fusion}: ASR/SE, vision, and ADS-B are fused before any advisory is issued.
All variants incorporate vision detections and ADS-B tracks within a common decision layer that outputs graded advisories to the pilot and/or controller.}

  \label{fig:topology}
\end{figure*}

\subsubsection{ASR for ATC/Pilot Communications}
Several automatic speech recognition (ASR) models have been proposed to reduce the Word Error Rate (WER) of ATC and pilot radio transmissions, with the aim of integration into ATC and pilot communication workflows. A comprehensive review of recent approaches can be found in the special collection by Helmke et al.~\cite{helmke_automatic_2024}. Notably, substantial improvements in WER have recently been achieved by fine-tuning OpenAI's Whisper ASR model~\cite{Whisper} on simulated and synthetic ATC speech datasets~\cite{van_doorn_whisper-atc_2024, garib_simugan-whisper-atc_nodate}.

Despite recent WER reductions, the safety impact remains uncertain. WER measures transcription accuracy, not how recognition errors influence pilot/controller performance during critical events. As noted by van Doorn et al.~\cite{van_doorn_applying_2023}, concrete performance requirements for safety management are not well specified. Moreover, the events where ASR would help most are rare and highly context dependent (airport geometry, traffic density, fatigue).

\subsubsection{Visual Scene Understanding}
Image processing advances in the ATC context include the identification of aircraft in the air~\cite{li_lightweight_2022,radovic_object_2017}, on the ground~\cite{pratama_system_2024, singh_deep_2024, alshaibani_airplane_2021}, or in operational contexts~\cite{kim_air_2021}. These models are often built on top of the YOLO object detection architecture~\cite{YOLO}.

\subsection{Architectural Topologies}
Assistive systems offer various design choices: placement of ASR/Speech enhancement (SE) relative to pilot audio (ASR/SE-first or parallel), fusion approach for vision and ADS-B (early or late), and recipient of advisories (pilot, controller, or both). Figure~\ref{fig:topology} illustrates three representative configurations, highlighting the need for flexible simulation environments to systematically evaluate diverse architectures across various operational scenarios.

\section{Challenges}\label{sec:challenges}
Evaluating air traffic situational awareness systems presents four key challenges: (1) moving from subtask accuracy to operational safety gains, (2) modeling rare failure scenarios with limited real-world data, (3) integrating multimodal context in evaluation frameworks, and (4) enabling simulation environments that support human-in-the-loop input and intelligent system responses.

\subsection{Operational Evaluation}
Subtask performance metrics, such as ASR’s WER, are insufficient alone to characterize safety system performance, since they do not capture how recognition errors interact with airspace geometry, traffic, and operator state. In AIRHILT we therefore emphasize scenario-based evaluation, using metrics such as time-to-first-warning and avoided conflicts to quantify the contribution of assistive systems to preventing unsafe outcomes.

\subsection{Human-in-the-Loop Operation}

A related challenge for the design of the simulation environment is to incorporate humans in the scenarios via simulated ATC and pilot workstations. To actually execute operational evaluations as described above, the environment must support pilot- and controller-in-the-loop operation, so that human behaviors can be incorporated into the chain of events.

\subsection{Scarcity of Failure Data}
Safety-critical events are rare and underrepresented in public datasets, making the evaluation of assistive systems for conflict detection challenging. Therefore, these systems typically require evaluation through carefully constructed simulated scenarios developed with expert input and supplemented by synthetic generation and real ATC data when available. However, effectively decreasing the sim-to-real gap between actual incidents and simulated scenarios remains a significant challenge.

\subsection{Multimodal Context Integration}

Current evaluations often rely on narrow, single-channel data inputs. Richer simulation frameworks should be able to synchronize visual, audio, and state data (e.g., from TartanAviation~\cite{dellarocco_tartanaviation_2025}) to reflect the true complexity of operational environments.

\section{Simulation Environment and Interfaces}\label{sec:env}

\begin{figure*}[t]
  \centering
  \includegraphics[width=\textwidth]{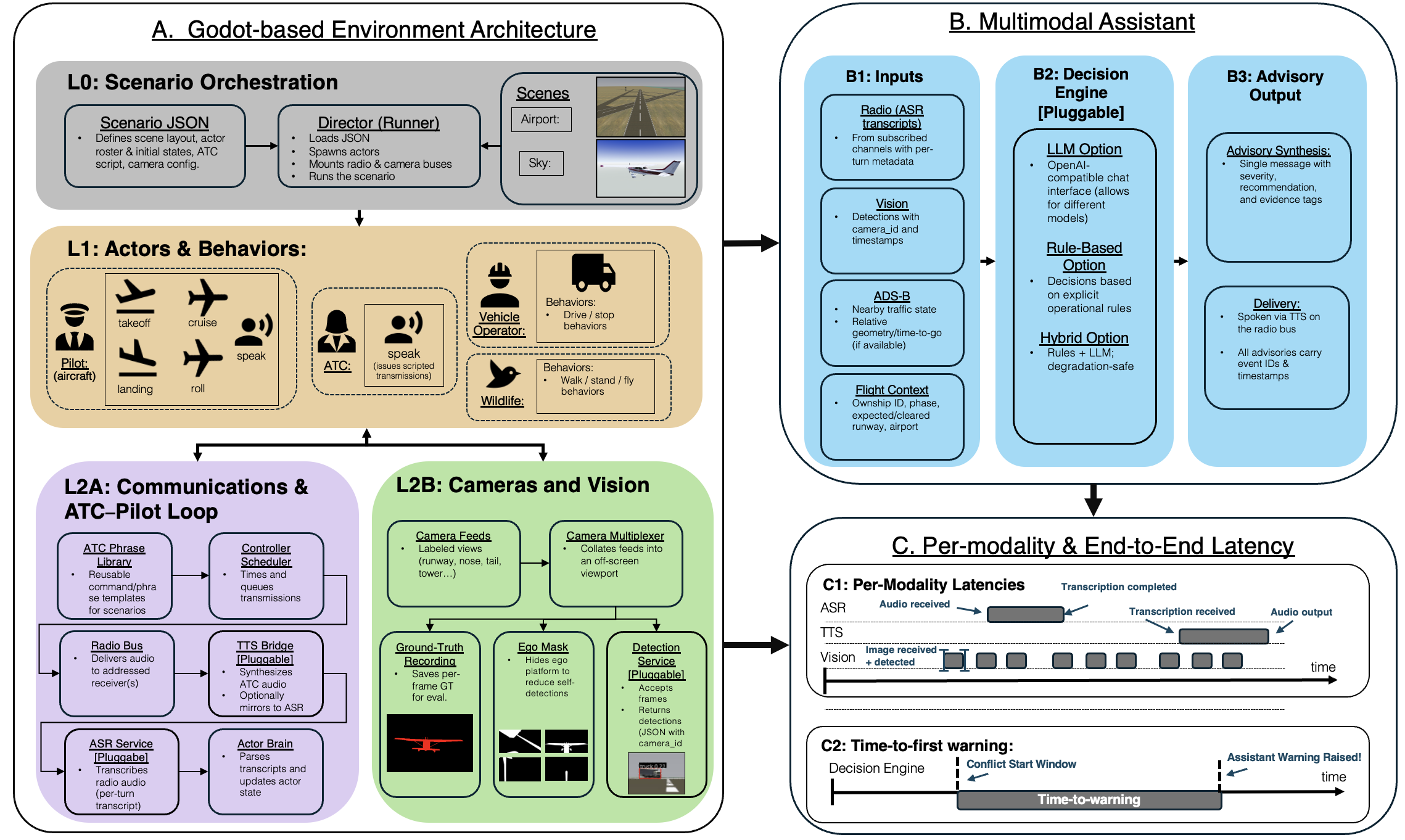}
  \caption{Modular simulation environment and onboard assistant architecture.
  A) Godot-based environment with scenario orchestration (L0), actors (L1), and I/O subsystems (L2A/L2B). 
  B) Multimodal assistant with pluggable decision engine and advisory output. 
  C) Built-in logging for per-modality and end-to-end latencies.}
  \label{fig:env_overview}
\end{figure*}

To enable efficient and scalable evaluation of different multimodal pilot-assist architectures with representative human-in-the-loop interactions, we introduce \emph{AIRHILT}, a lightweight, modular simulation environment built upon the Godot engine. \emph{AIRHILT} offers a unified simulation platform that synchronizes pilot-to-ATC communications, camera-based visual perception, and ADS-B traffic data, enabling systematic evaluation of assistive systems designed for aviation conflict detection and resolution.

\subsection{Design Goals and Scope}

The design of \emph{AIRHILT} directly targets the operational and methodological challenges outlined in Section~\ref{sec:challenges}, structured around the following primary design goals:

\subsubsection{Modularity and Interoperability}
All components communicate via stable REST/JSON interfaces, allowing ASR, vision, decision, and text-to-speech (TTS) modules to be swapped with minimal code changes.

\subsubsection{Reproducibility}
Each simulation run leverages deterministic seeding, stable event identifiers, and unified timestamps, enabling consistent and repeatable experimentation.

\subsubsection{Representative Human-in-the-loop Interactions}
Frequency-specific, addressed radio traffic and confidence-gated actor behaviors approximate operational interactions.

\subsubsection{Logging and Timing}
Built‑in logging records per‑modality and end‑to‑end latencies for each advisory event.

\subsubsection{Deployment}
A lightweight Godot runtime with compact FastAPI services runs on a single workstation and benefits from optional GPU acceleration.

Fig.~\ref{fig:env_overview} presents a high-level overview of the \emph{AIRHILT} architecture, while the following subsections provide a deeper exploration into the design, functionality, and implementation details of each layer and component.

\subsection{Scenario Orchestration (L0)}

The scenario execution in \emph{AIRHILT} is orchestrated through a clearly structured, declarative \texttt{Scenario JSON}, providing researchers with full control over simulation initialization and runtime conditions. Each scenario specification encapsulates:

\begin{itemize} \item \textbf{Scene type and geometry:} Defines whether the scenario involves airport surface operations or en route airborne interactions, alongside geometric details for each actor's initial placement and orientation. \item \textbf{Actor roster and initial states:} Specifies actors such as aircraft, ATCs, vehicles, and wildlife, assigning each with initial behavior states. \item \textbf{Scripted ATC communication timeline:} Specifies structured, timestamped radio exchanges with unique identifiers, including addressed clearances and instructions, with options to inject controlled noise artifacts. \item \textbf{Camera configuration:} Determines the placement and orientation of cameras (e.g., aircraft-mounted and/or tower-based), along with sampling rates for visual perception components. \item \textbf{Randomization seed:} Controls deterministic variants for reproducible small geometric and timing perturbations, for example small shifts in clearance timing or initial longitudinal spacing between aircraft.
\end{itemize}

Upon ingestion of the Scenario JSON, the \texttt{Director (Runner)} automatically loads and initializes all scene assets and actors as defined in the scenario specification. It subsequently mounts synchronized communication and vision buses to maintain consistent alignment of audio (radio), visual (camera), and corresponding ground-truth data streams.

The environment explicitly supports two primary scene types selected for their operational relevance to aviation safety:

\begin{itemize}\RaggedRight
  \item \textbf{Airport Surface Scene:} Inspired by Ronald Reagan Washington National Airport (DCA), featuring intersecting runways and high‑complexity ground operations representative of common conflict scenarios.
  \item \textbf{En route Airspace Scene:} Represents typical airborne interactions, including converging flight paths and altitude‑based conflict geometries.
\end{itemize}

\subsection{Actors and Behavior Abstractions (L1)}

The environment defines a set of actor types and structured behaviors that aim to closely represent operational interactions in aviation scenarios, balancing fidelity and computational efficiency. These models support pilot‑in‑the‑loop interactions without the overhead of full‑fidelity flight dynamics. Additionally, the environment was intentionally designed to simplify the addition of new actor types and behaviors, with clear guidance provided in the repository documentation.

\begin{itemize}
  \item \textbf{Actor Types:} The environment currently models four primary actor classes:
  \begin{itemize}
    \item \emph{Pilot/Aircraft}: Capable of executing named behaviors such as takeoff, cruise, landing, and responding to radio commands.
    \item \emph{Air Traffic Controller (ATC)}: Issues scripted, addressed instructions and clearances, following structured timelines.
    \item \emph{Vehicle Operators}: Simulate ground vehicles with pre-defined driving patterns (e.g., drive, stop).
    \item \emph{Wildlife}: Incorporates animal actors exhibiting basic behaviors (walk, stand, fly), serving as hazard sources.
  \end{itemize}

 \item \textbf{Control Inputs}: For all non‑wildlife actors, addressed
radio transmissions are parsed with a slot-based method that approximates how real operators extract intent from speech (callsign, runway/altitude assignments, temporal instructions such as ``hold short'' or ``cleared for takeoff''). For example, ``N123AB, cleared for takeoff runway one nine'' is parsed into slots \texttt{callsign = N123AB}, \texttt{action = cleared\_for\_takeoff},
\texttt{runway = 19}. Low‑confidence or ambiguous inputs trigger structured clarification behavior at the actor layer.

  \item \textbf{Physics-based Motion}: Actor movements use computationally efficient, physics-informed models sufficient to maintain timing and geometric realism in conflict scenarios. Motion behaviors include timed vertical maneuvers (climbs and descents), and relatively simplified representations of landing (glide, flare, rollout) and takeoff (roll, rotate, climb-out) phases, among others. Detailed equations and parameter choices are provided in the repository.
\end{itemize}

\subsection{Communications Loop: Radio, TTS, and ASR (L2A)}

The communications subsystem of \emph{AIRHILT} is structured to approximate operational aviation radio interactions while supporting controlled experimentation with pilot and ATC communications. This communication loop incorporates clearly defined interfaces for radio transmissions, TTS synthesis, and ASR, all implemented as interchangeable modules accessible through standardized REST/JSON endpoints:

\begin{itemize}

\item \textbf{Radio Bus}: The radio subsystem utilizes frequency-specific channels for addressed transmissions and includes optional overhearing capabilities. Transmission guarantees include ordered delivery, stable identifiers for each radio turn, and precise emission timestamps (\(t_{\text{tx}}\)).

\item \textbf{ASR Service (Interchangeable)}: Pluggable ASR models transcribe received audio and emit finalization timestamps (\(t_{\text{asr\_out}}\)) and optional confidences. By default, the ASR runs in parallel to the pilot audio path (matching the topology in Fig.~\ref{fig:topology}b), so pilots hear raw audio while a copy is forwarded to the recognizer; users may alternatively configure the ASR output as an intermediate step prior to pilot reception, depending on the assistive architecture and experimental setup being evaluated.

\item \textbf{TTS Bridge (Interchangeable)}: The environment integrates pluggable TTS services (e.g., XTTS-v2~\cite{noauthor_coqui_nodate}) that convert scripted text instructions into realistic audio. An optional augmentation layer adds configurable radio-channel effects, including noise and distortion, with precise control over signal-to-noise ratio (SNR) and mixing parameters. In our experiments, all radio-path audio is encoded as single-channel PCM at a $16\,\text{kHz}$ sample rate to match the reference ASR and SimuGAN configurations.

\end{itemize}

To support latency evaluation, the logger records the timestamps needed to compute per‑module and end‑to‑end timings (e.g., \(t_{\text{tx}}\), \(t_{\text{asr\_out}}\), augmentation/SNR settings, and stable transmission IDs).

\subsection{Camera and Vision Pipeline (L2B)}

The visual perception subsystem of \emph{AIRHILT} provides configurable, synchronized camera feeds designed to replicate representative visual data streams. This pipeline maintains clearly defined and stable REST/JSON interfaces for seamless module interchangeability, supporting systematic experimentation and reproducibility:

\begin{itemize}

\item \textbf{Camera Feeds}: The environment includes multiple predefined camera perspectives such as runway views, aircraft nose and tail cameras, and tower-based views. These feeds are captured at configurable sampling rates (typically $16$--$20\,\mathrm{Hz}$ in our reference configuration) and resolutions ($W \times H$ pixels), with optional per-actor sampling control for targeted experimentation.

\item \textbf{Multiplexer}: A configurable multiplexer mirrors camera feeds to off-screen viewports, enabling simultaneous data capture without disrupting primary simulation rendering performance.

\item \textbf{Ego Mask}: Per-camera visibility masks are implemented to suppress model self-detections.

\item \textbf{Detection Service (Interchangeable)}: Visual detection services operate as independently swappable modules accessible via REST APIs. They receive images (alongside optional ego masks), timestamps (\texttt{ts\_ms}), and camera identifiers (\texttt{camera\_id}), and return structured detections in JSON. Each detection object includes classification labels, confidence scores, and bounding box coordinates.

\item \textbf{Ground Truth Data}: Each actor is assigned a unique color in dedicated viewport renders. From these per‑frame masks we derive presence labels and ideal (axis‑aligned) bounding boxes, and log them alongside class IDs for analysis and visualization.

\end{itemize}

\afterpage{%
\begin{figure*}[!t]
  \centering
  \setlength{\tabcolsep}{0pt}%
  \renewcommand{\arraystretch}{0}%
  \setlength{\fboxsep}{0pt}%
  \setlength{\fboxrule}{0.25pt}%

  \begin{tabular}{@{}ccc@{}}
    \begin{minipage}[t]{.333\textwidth}\centering
      {\footnotesize (a) S01A Runway overlap\par}
      \fbox{\includegraphics[width=\linewidth]{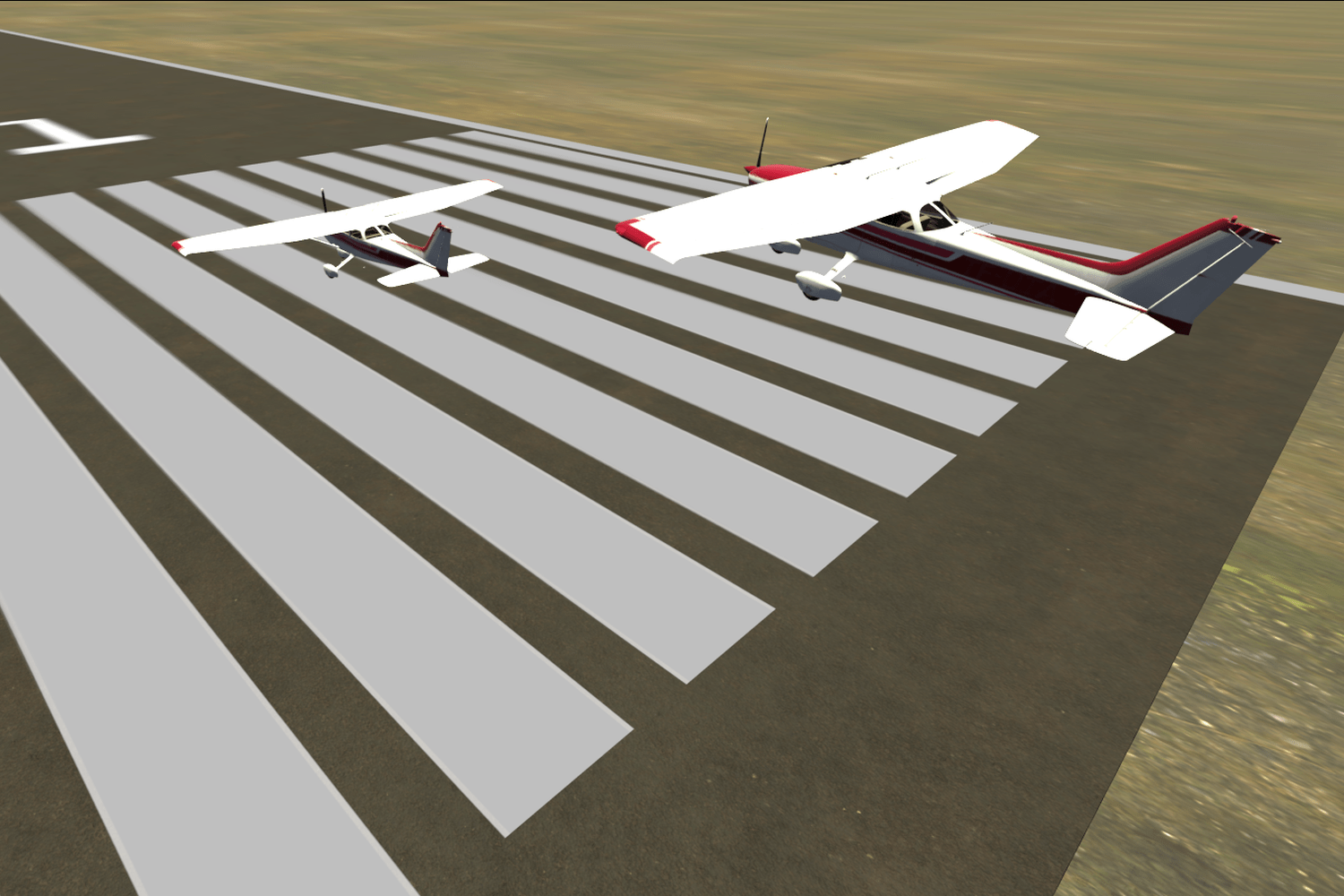}}
    \end{minipage}
    &
    \begin{minipage}[t]{.333\textwidth}\centering
      {\footnotesize (b) S01B Vehicle incursion\par}
      \fbox{\includegraphics[width=\linewidth]{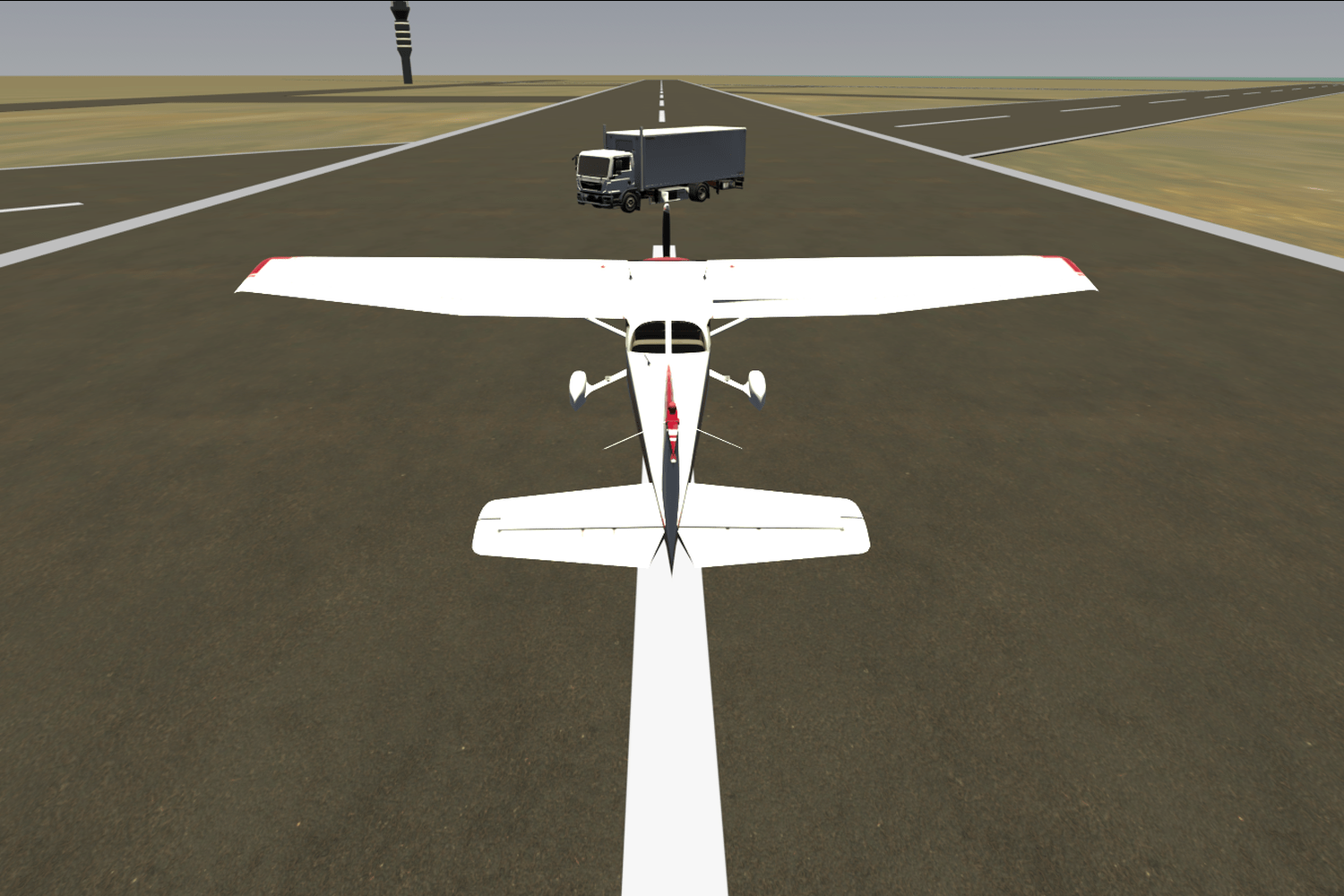}}
    \end{minipage}
    &
    \begin{minipage}[t]{.333\textwidth}\centering
      {\footnotesize (c) S01C Wildlife incursion\par}
      \fbox{\includegraphics[width=\linewidth]{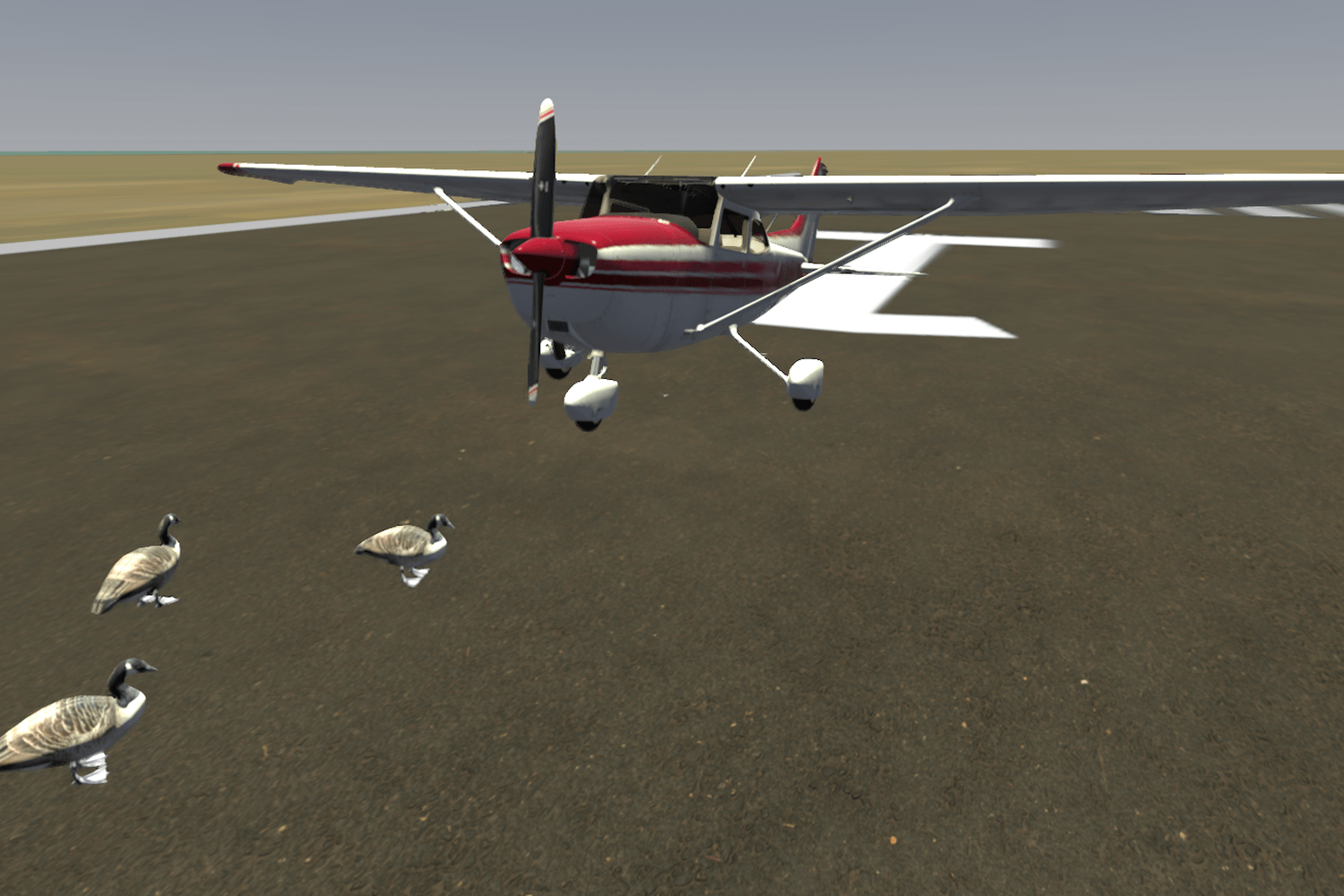}}
    \end{minipage}
    \\[-0.35ex]
    \begin{minipage}[b]{.333\textwidth}\centering
      \fbox{\includegraphics[width=\linewidth]{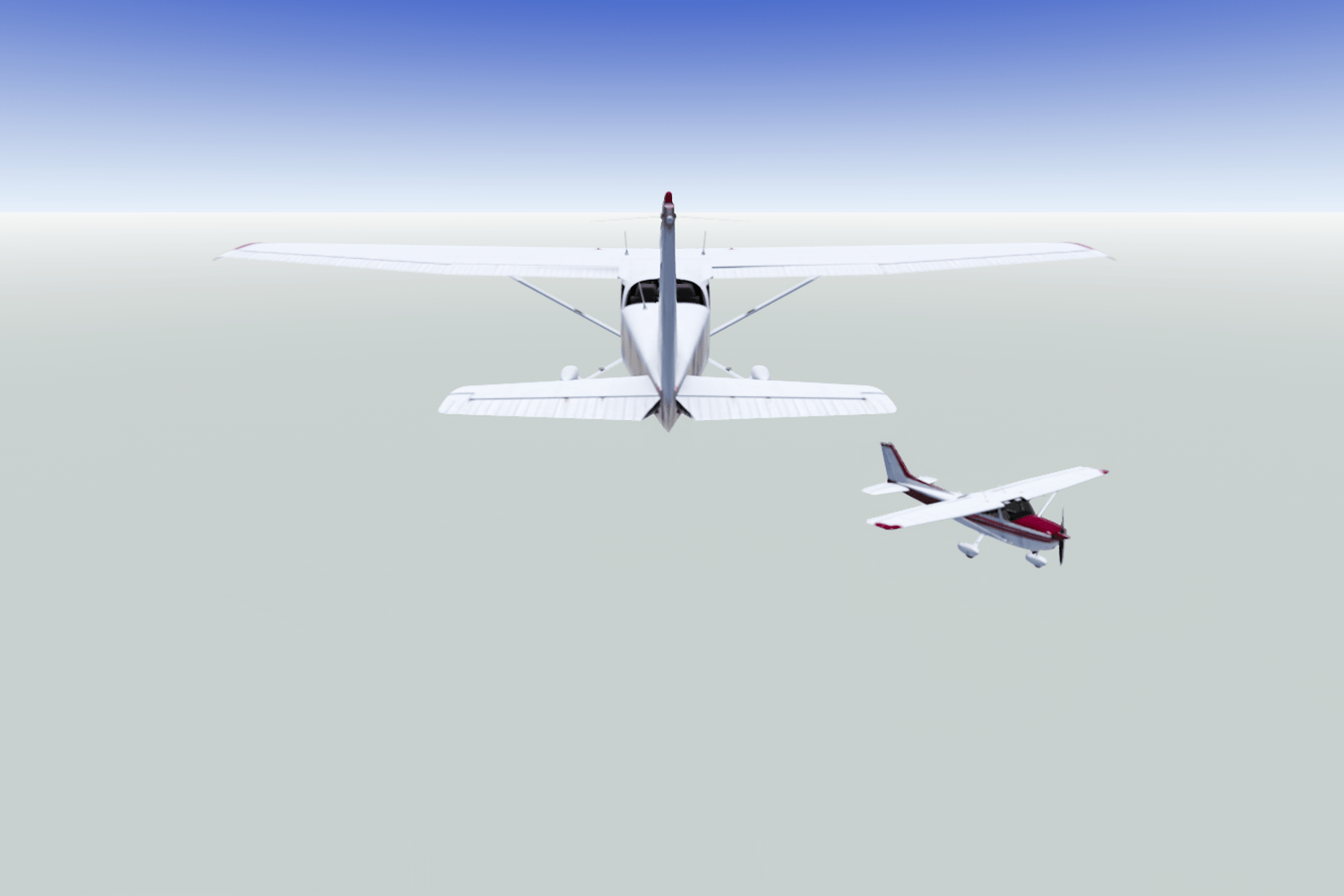}}\\[-0.25ex]
      {\footnotesize (d) S02A Geometric conflicts}
    \end{minipage}
    &
    \begin{minipage}[b]{.333\textwidth}\centering
      \fbox{\includegraphics[width=\linewidth]{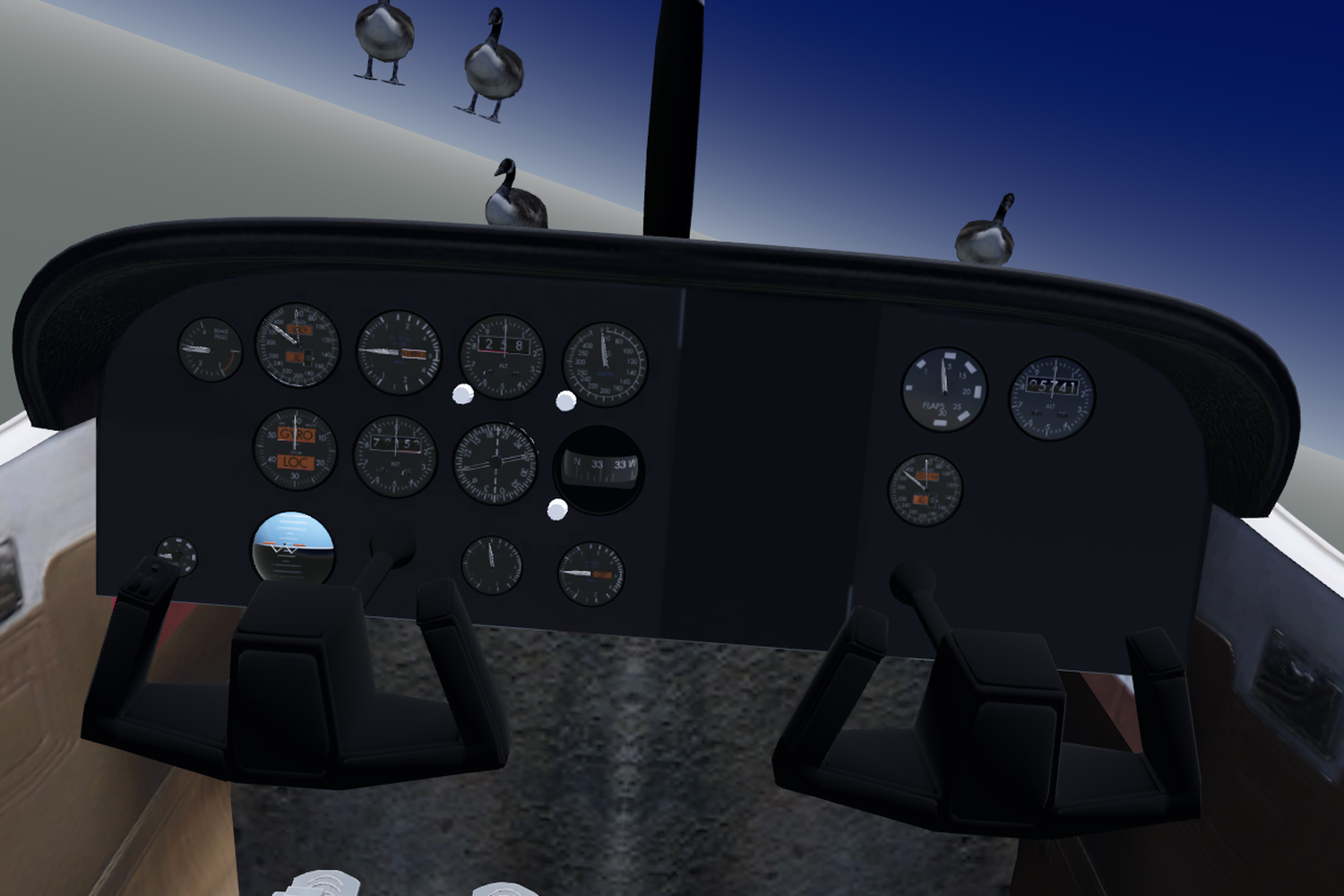}}\\[-0.25ex]
      {\footnotesize (e) S02B Airborne emergencies}
    \end{minipage}
    &
    \begin{minipage}[b]{.333\textwidth}\centering
      \fbox{\includegraphics[width=\linewidth]{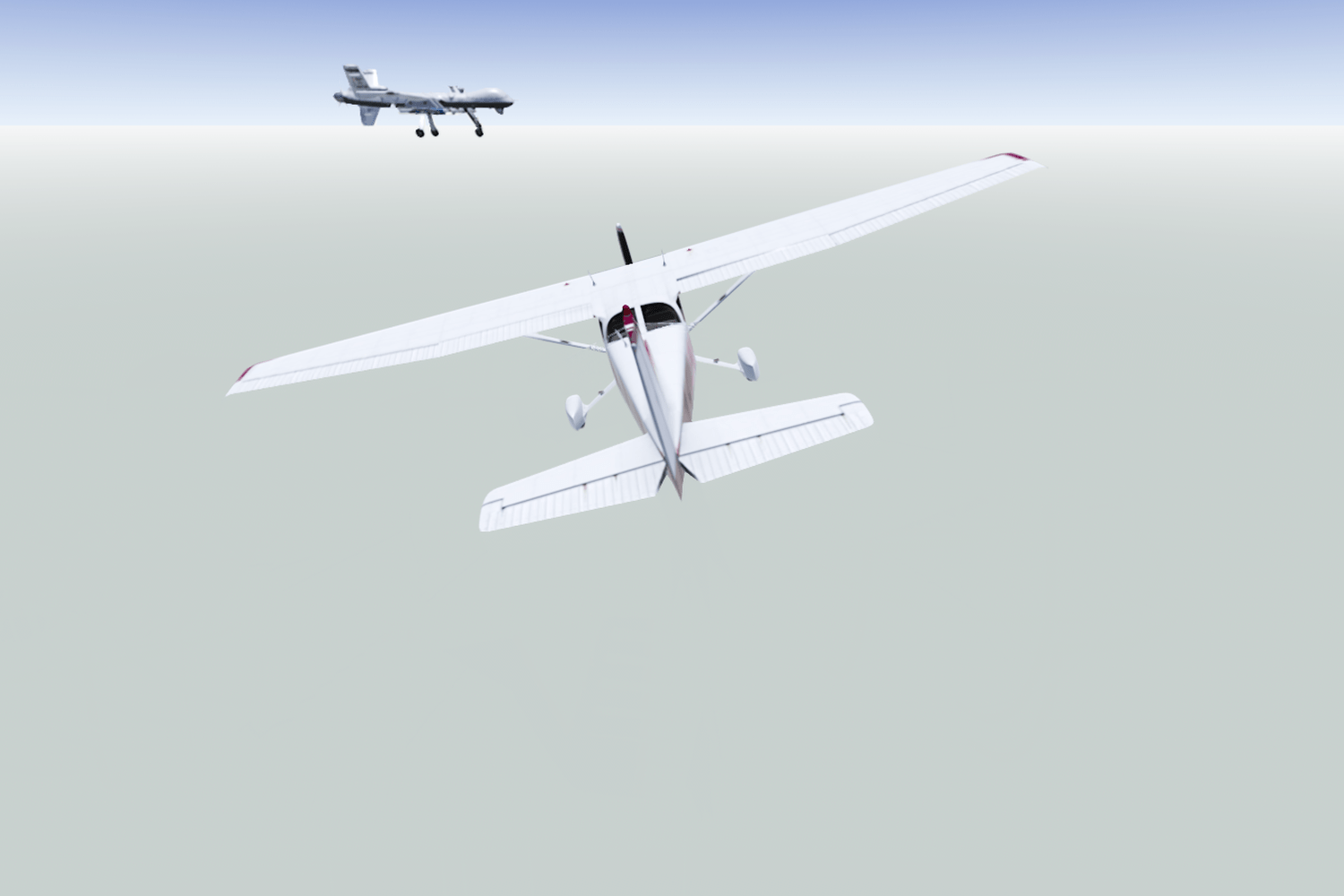}}\\[-0.25ex]
      {\footnotesize (f) S02C Non-cooperative intruders}
    \end{minipage}
  \end{tabular}

  \caption{Representative examples from the six scenario families used for evaluation.
  Top row shows terminal-area scenarios (runway overlaps S01A, vehicle incursions S01B, wildlife incursions S01C).
  Bottom row shows en route scenarios (geometric conflicts S02A, airborne emergencies S02B, non-cooperative intruder encounters S02C).}
  \label{fig:scenario_examples}
\end{figure*}
} 

\subsection{Multimodal Assistant Interface (B): Inputs, Decision Engine (Interchangeable), and Outputs}

The multimodal assistant exposes a well‑specified interface that defines inputs, outputs, and the pluggable decision module. This decouples environment integration from any particular algorithm and allows alternative implementations to be swapped with minimal code changes elsewhere.

\begin{itemize}

\item \textbf{Inputs}: The assistant receives synchronized multimodal data in three categories:
\begin{itemize}
\item \emph{Radio Transcripts}: Structured sequences of communications with timestamps (\texttt{ts\_ms}), speaker IDs, transcript text, frequency, unique turn IDs, optional \emph{overhearing/subscription} flags (to receive nearby actors’ traffic), addressed callsigns, and confidences when available.

\item \emph{Vision Detections}: Lists of detection results from visual perception, each containing frame timestamps (\texttt{ts\_ms}), associated camera identifiers, detected object classifications, confidence scores, and bounding box coordinates.
\item \emph{ADS-B and Flight Context}: Data slices summarizing ownship operational states, expected and cleared runway information, and positional and velocity tracks.
\end{itemize}

\item \textbf{Decision Engine (Interchangeable)}: Decision-making modules are integrated via standardized HTTP/JSON interfaces, allowing simple substitution between the different options of rule-based logic, LLM-based reasoning, or other hybrid methods. Inputs are posted as structured requests, and modules return standardized advisory objects (message, severity, optional recommendations, metadata).

\item \textbf{Outputs}: The output schema is user‑configurable. Advisories could include concise text, a severity level (INFO/ADVISORY/CAUTION/WARNING), optional recommendations, and supporting metadata for traceability and debugging.

\item \textbf{Severity scale and speech threshold}: In the reference pipeline we use a four-level severity scale \{INFO, ADVISORY, CAUTION, WARNING\} and map these to integer levels; only advisories at or above a configurable threshold \texttt{SPEAK\_MIN\_LEVEL} are synthesized via TTS (Section~\ref{sec:reference_pipeline}), so that lower-severity findings can be logged without contributing to pilot workload.

\item \textbf{Advisory Delivery}: Advisory objects are synthesized into audible alerts via the previously described TTS subsystem and delivered through the appropriate radio channel.
\end{itemize}

\subsection{Timebase and Latency Accounting (C)}
We instrument all subsystems with a shared monotonic timebase and log start/end events for each module. For radio/ASR we record the transmission time \(t_{\text{tx}}\) and ASR finalization \(t_{\text{asr}}\). For vision we record frame exposure end \(t_{\text{frame}}\) and detector completion \(t_{\text{vision}}\). For ADS\mbox{-}B we record ingest \(t_{\text{adsb,in}}\) and post‑processor output \(t_{\text{adsb,out}}\). The decision engine records when all required inputs are available \(t_{\text{ready}}\) and when an advisory is produced \(t_{\text{dec}}\) (this includes LLM inference when used). The audio path records the first audible sample delivered to the radio bus \(t_{\text{tts}}\). Each scenario annotates the conflict‑window opening time \(t_{\text{conflict}}\).

From these timestamps we compute per‑module and end‑to‑end timings during post‑processing:
\begin{align}
\text{ASR latency} &= t_{\text{asr,out}} - t_{\text{tx}}, \label{eq:asr_latency}\\
\text{Vision latency} &= t_{\text{vision}} - t_{\text{frame}}, \label{eq:vision_latency}\\
\text{ADS\mbox{-}B latency} &= t_{\text{adsb,out}} - t_{\text{adsb,in}}, \label{eq:adsb_latency}\\
\text{Decision latency} &= t_{\text{dec}} - t_{\text{ready}}, \label{eq:decision_latency}\\
\text{TTS latency} &= t_{\text{tts}} - t_{\text{dec}}, \label{eq:tts_latency}\\
\text{Time-to-first-warning} &= t_{\text{tts}} - t_{\text{conflict}}. \label{eq:ttfw}
\end{align}

All timestamps share a common monotonic timebase within the simulation process, which allows these per-module and end-to-end latencies to be compared across runs and across alternative assistant implementations. Figure~\ref{fig:env_overview}C illustrates the latencies and warning intervals.

\section{Evaluation Scenarios} \label{sec:scenarios}
Six primary scenario families were developed to evaluate diverse, realistic aviation conflicts across terminal and en route airspace, illustrated in Figure~\ref{fig:scenario_examples}. We focus exclusively on human-, procedural-, and communication-driven conflicts, whereas mechanical failures are out of scope in the current release.

In terminal airspace (airport surface operations), three scenario families assess runway incursions and occupancy conflicts:

\begin{itemize}\item \textbf{S01A (Runway Overlap)} evaluates conflicts such as miscommunication during runway clearances, including bad readbacks, missed cancellation transmissions, and misaddressed instructions. \item \textbf{S01B (Vehicle Runway Incursions)} involves ground vehicle incursions due to delayed or dropped HOLD commands, misaddressing, or intentional noncompliance. \item \textbf{S01C (Wildlife Runway Incursions)} addresses wildlife presence hazards, particularly delayed wildlife warnings or failures to detect wildlife incursions. \end{itemize}

In en route airspace (sky domain), three scenario families cover airspace geometry and coordination conflicts:

\begin{itemize} \item \textbf{S02A (Geometric Airspace Conflicts)} evaluates situations like in-trail closing, head-on trajectories, and vertical separation violations. \item \textbf{S02B (Airborne Emergency Coordination)} assesses emergency situations such as bird strikes and engine-out drift-down scenarios requiring precise coordination. \item \textbf{S02C (Uncoordinated Intruders)} focuses on encounters with non-cooperative aircraft lacking ADS-B/TCAS coordination, requiring purely visual detection. \end{itemize}

The environment was deliberately designed to simplify the creation of these scenarios, providing a structured and intuitive JSON-based configuration. A comprehensive guide to scenario creation, detailing workflows, parameters, and extensions, is included in our public repository.

\section{Reference Pipeline}
\label{sec:reference_pipeline}

To demonstrate \textit{AIRHILT}, we implemented a reference pipeline and tested it on \textbf{S01A---Runway Overlap}. For instance, in the \emph{bad readback accepted} variant, an arrival is cleared to land runway~01, the pilot incorrectly reads back runway~19, the tower replies ``roger,'' and later a departure is cleared for takeoff on runway~01, creating an occupancy conflict. These variants allow us to study end-to-end assistant behavior at the scenario level, in terms of whether and when warnings are raised relative to the opening of the conflict window. Other S01A conflict types exercised in the suite include: \emph{cancel takeoff not received} (tower cancels the departure but the intended aircraft never hears it), \emph{misaddressed takeoff clearance} (clearance spoken with the wrong callsign, departure accepts), and \emph{tight timing overlap} (both clearances valid, but spacing is insufficient).

\subsection{Models and Interfaces}
Table~\ref{tab:design_choices} summarizes the modules and essential information regarding the models used in the reference pipeline.

\begin{table*}[t]
\centering
\footnotesize
\begin{threeparttable}
\caption{Radio‑path realism via SimuGAN (approx.\ 4.5~h TartanAviation ATC) or DSP, applied only to ATC--pilot comms (WAV/TTS); advisory TTS is not noise‑augmented, and GPT-OSS-20B is used only for advisory text surface form. See the public repository for further implementation details.}
\label{tab:design_choices}
{\setlength{\tabcolsep}{3.4pt}\renewcommand{\arraystretch}{1.14}
\begin{tabularx}{\textwidth}{@{} l P{0.19\textwidth} P{0.21\textwidth} Y @{}}
\toprule
\textbf{Subsystem} & \textbf{Model / version} & \textbf{Training / fine-tune} & \textbf{Key I/O and behavior} \\
\midrule
ASR & Whisper (OpenAI), fine-tuned & ATC-focused fine-tune (\texttt{large-v2})~\cite{garib_simugan-whisper-atc_nodate} &
\makecell[l]{Inputs: speaker, frequency.\\ Output: transcript and $t_{\mathrm{ASR}}$.\\ Notes: runway token canonicalization;\\ confidence gate $\ge \tau_{\mathrm{ASR}}$.} \\
\addlinespace[3pt]
TTS (advisories) & Coqui XTTS-v2~\cite{noauthor_coqui_nodate} & --- &
\makecell[l]{Advisory audio (no noise augmentation).} \\
\addlinespace[3pt]
Radio noise (ATC--pilot) & \textbf{SimuGAN} (learned RF/VHF); DSP fallback (SNR) &
approx.\ 4.5\,h TartanAviation ATC (for SimuGAN)~\cite{garib_simugan-whisper-atc_nodate} &
\makecell[l]{Applied only to ATC and pilot received audio.\\ Controls: SNR, wet mix, profile.} \\
\addlinespace[3pt]
Vision & Ultralytics YOLOv10~\cite{wang_yolov10_2024}; ego-masking; tiled inference &
Class filter (airplane, truck, bird) &
\makecell[l]{Outputs: detections and $t_{\mathrm{vision}}$.\\ Multi-camera corroboration ($K$) or confidence persistence.} \\
\addlinespace[3pt]
ADS-B / flight context & Parser + roster filters (airport map) & --- &
\makecell[l]{Outputs: runway expectations, occupancy, and tracks.} \\
\addlinespace[3pt]
Decision engine & Rules + GPT-OSS-20B (natural language generation, NLG only) & --- &
\makecell[l]{Rule ladder; occupancy + time-to-go (TTG);\\
see Alg.~\ref{alg:decision} for details.} \\
\addlinespace[3pt]
Output object & Advisory object + delivery & --- &
\makecell[l]{Fields: severity, message, recipients, evidence;\\
speak only if $\ge$ {\ttfamily\small SPEAK\_\allowbreak MIN\_\allowbreak LEVEL}.} \\
\bottomrule
\end{tabularx}}
\end{threeparttable}
\end{table*}

\subsection{Decision Logic}
Algorithm~\ref{alg:decision} outlines how the decision engine integrates information from the different modalities to determine when and what warnings to raise.

In the evidence-fallback branch we compute a scalar score
\(S = 0.50\,W_V + 0.35\,W_A + 0.15\,W_C\), where \(W_V\), \(W_A\), and \(W_C\) are normalized evidence terms derived from vision occupancy, ASR consistency, and overall conflict context (all in \([0,1]\)). In our S01A experiments we set \(\tau_{\mathrm{ASR}} = 0.8\) and \(\tau_{\mathrm{vis}} = 0.7\), which were chosen empirically to balance missed detections and nuisance alerts.

\begin{algorithm}[!t]
\caption{Decision engine for S01A family (parallel modality logic)}
\label{alg:decision}
\KwInput{Radio turns $\mathcal{A}$; vision detections $\mathcal{V}$; ADS\mbox{-}B/roster $\mathcal{B}$; thresholds $\tau_{\mathrm{ASR}},\tau_{\mathrm{vis}}$}
\KwOutput{Advisory object $\langle$type, severity, text, evidence$\rangle$}
\BlankLine

\textbf{Modality updates (run in parallel):}
\begin{enumerate*}[label=(\roman*)]
\item \emph{ASR parse:} normalize runway tokens; compute \textit{slot\_conf}.
\item \emph{Vision occupancy:} multi‑cam corroboration ($K$) or conf‑persistence; produce activity/occupancy flags.
\item \emph{ADS‑B slice:} compute TTG and determine nearby traffic.
\end{enumerate*}\;

\textbf{Guard:} if \textit{slot\_conf}$<\tau_{\mathrm{ASR}}$ then \textbf{request clarification}.

\textbf{On any stream update, evaluate ladder gates:}
\begin{enumerate*}[label=\alph*)]
\item readback mismatch $+\,$activity $\Rightarrow$ \textbf{CAUTION};
\item occupancy $+$ (TTG$\le$\,8\,s or arrival context) $\Rightarrow$ \textbf{WARNING};
\item recipient ambiguity $+\,$activity $\Rightarrow$ \textbf{CAUTION};
\end{enumerate*}\;

\textbf{If none fired (evidence fallback):} score $S \leftarrow 0.50\,W_V + 0.35\,W_A + 0.15\,W_C$; if $S\!\ge\!0.75$ $\Rightarrow$ \textbf{CAUTION}; if $S\!\ge\!0.50$ $\Rightarrow$ \textbf{ADVISORY}\;

\textbf{Compose advisory text} from rules (optionally reformulated by GPT‑OSS‑20B) and \textbf{deliver via TTS}\;

\textbf{Return} advisory object with evidence (radio IDs, camera IDs, TTG, rules triggered)\;
\end{algorithm}

This ladder of rules prioritizes fast escalation in clearly unsafe conditions (for example, conflicting clearances with observed runway activity), while the evidence score \(S\) provides a conservative fallback mechanism in more ambiguous cases.

\subsection{Preliminary Results}
We evaluated \textbf{S01A} across 10 runs per conflict type with randomized visibility and SimuGAN SNR settings to stress both the vision and radio paths. Averages across these runs were: time-to-first-warning \(t_{\mathrm{tts}}-t_{\mathrm{conflict}} \approx \textbf{7.66\,s}\); ASR latency \(t_{\mathrm{asr}}-t_{\mathrm{tx}} \approx \textbf{5.88\,s}\); vision latency \(t_{\mathrm{vision}}-t_{\mathrm{frame}} \approx \textbf{0.415\,s}\); and \textbf{TTS synthesis/delivery} \(t_{\mathrm{tts}}-t_{\mathrm{dec}} \approx \textbf{0.9\,s}\). The first vision detection typically occurred at $\sim 125\,\text{m}$ range. These results show that, even with realistic radio noise and degraded visibility, the assistant can deliver runway-overlap warnings several seconds after the conflict window opens.

In‑depth results across all scenario families will be made available in the public repository.

\section{Concluding Remarks}\label{sec:limitations}
\emph{AIRHILT} provides an open-source and flexible environment that facilitates experimentation with different multimodal assistive systems in aviation. The present evaluation focuses on the S01A runway-overlap scenarios. Extending quantitative assessment to the remaining scenario families is an important direction for future work, and updated results will be reported in the public repository.

\subsection{Limitations}
Current limitations of \emph{AIRHILT} include a sim-to-real gap, particularly within the vision detection pipeline, where simulation artifacts can lead to discrepancies from real-world performance. Human-factors validation is also limited, as we have not yet conducted controlled pilot and ATC studies to quantify workload, situational awareness, and operator acceptance of the multimodal assistant. In addition, the reference assistant implementation is not yet latency-optimized; our goal in this work is to establish a clear baseline and measurement framework rather than to minimize processing time.

\subsection{Avenues for Future Work}
Future efforts will focus on simplifying environment setup procedures and improving the clarity and flexibility of the provided interfaces, aiming to reduce onboarding time and support rapid experimentation. Additionally, we plan to conduct simulator studies involving pilots and controllers to improve the fidelity and realism of modeling human-in-the-loop interactions within AIRHILT.

\newpage
\bibliographystyle{ieeetr}
\bibliography{references}

\end{document}